\documentclass[12pt,leqno]{article}
\usepackage{latexsym}
\newtheorem{theorem}{Theorem}

\newtheorem{lemma}{Lemma}

\newcommand{\blackslug}{\mbox{\hskip 1pt \vrule width 4pt height 8pt 
depth 1.5pt \hskip 1pt}}
\newcommand{\QED}{\quad\blackslug\lower 8.5pt\null\par\noindent}
\newcommand{\proof}{\par\penalty-100\vskip .5 pt\noindent{\bf Proof\/: }}
\newcommand{\ru}{\rule[-0.4mm]{.1mm}{3mm}}
\newcommand{\nni}{\ru\hspace{-3.5pt}}

\newcommand{\NI}{\mbox{$\: \nni\sim$}}

\newcommand{\cC}{\mbox{${\cal C}$}}

\newcommand{\cL}{\mbox{${\cal L}$}}
\newcommand{\cM}{\mbox{${\cal M}$}}

\newcommand{\Cn}{\mbox{${\cal C}n$}}

\newcommand{\subseteqf}{\mbox{$\subseteq_{f}$}}

\title{The logical meaning of Expansion}
\author{Daniel Lehmann\thanks{This work was done while the author was
on sabbatical leave at the Robotics Laboratory, Stanford University}
\\Institute of Computer Science, \\Hebrew University, \\Jerusalem 91904, Israel
\\lehmann@cs.huji.ac.il
}
\date{August 6th, 1999}
\begin{document}
\parindent0.0cm
\maketitle
\begin{abstract}
The Expansion property considered by researchers in Social Choice
is shown to correspond to a logical property of nonmonotonic
consequence relations that is the {\em pure}, i.e., not involving
connectives, version of a previously known weak rationality condition.
The assumption that the union of two definable sets of models is definable
is needed
for the soundness part of the result.
\end{abstract}
\section{Introduction}
In previous work~\cite{LehLogSemantics:TR}, 
I have shown that a set of three properties
of choice functions, previously studied by researchers
in the theory of social preferences: 
Contraction, Coherence and Local Monotonicity corresponds exactly to
an important family of nonmonotonic consequence operations, characterized
by five properties: Inclusion, Idempotence, Cautious Monotonicity,
Conditional Monotonicity and Threshold Monotonicity.
The choice functions that, in addition, satisfy a condition proposed
by Arrow correspond exactly to the operations that, in addition,
satisfy Rational Monotonicity.

Another property has been widely considered in Social Choice:
\[
{\bf Expansion} \ \ f(X) \cap f(Y) \subseteq f(X \cup Y).
\] 
It does not follow from Contraction, Coherence and Local Monotonicity.
Under the simplifying assumption that all sets are definable,
Expansion follows from Contraction and Arrow.

It is therefore a reasonable assumption that Expansion 
(in the presence of Contraction, Coherence and Local Monotonicity),
corresponds exactly
to some property of the consequence operation that is weaker than
Rational Monotonicity. Many such properties have been studied
in~\cite{KLMAI:89,Satoh:90,FLMo:91,Freund:InjDisj,FrLeh:NegRat}.

It shall be shown that, indeed, Expansion is, essentially, 
the exact equivalent of one of those properties: the property found 
by Satoh~\cite{Satoh:90}
and Freund~\cite{Freund:InjDisj} to characterize the existence 
of injective preferential models,
typically written, in the finitary framework:
if \mbox{$a \vee b \NI c$}, then there exists $a'$ and $b'$ such that
\mbox{$a \NI a'$}, \mbox{$b \NI b'$} and \mbox{$a' \wedge b' \models c$},
where $\models$ is logical implication.
In the infinitary framework, the same property is: 
\begin{equation}
\label{eq:satoh} 
\cC(\Cn(A) \cap \Cn(B)) \subseteq \Cn(\cC(A) , \cC(B)).
\end{equation}

Two important remarks must be made immediately.
The property above assumes the existence of an underlying
monotonic consequence operation ( $\models$, \Cn), 
of which \cC\ is an extension. 
We do not assume the existence of such an operation and therefore,
shall use, in place of $\Cn(A)$, the closest operation that can be defined
in terms of \cC, i.e.,
\mbox{$\bigcap_{F \supseteq A} \cC(F)$}.
The finitary form also         
assumes the existence of a disjunction
in the language. We shall need to make a similar assumption:
we shall assume that the union of two definable sets is definable.
The fact that such an assumption is needed indicates that
Expansion is not a property that is as natural as the other 
properties considered.
The question of whether the result holds without
this assumption is open.

The property we propose to consider is:
\[
{\bf (E)} \ \ \ \ \ \cC(\bigcap_{F \supseteq A {\rm \ or \ } F \supseteq B} \cC(F)) \:\subseteq \: \cC(\cC(A) , \cC(B)).
\]
Notice that, due to the Threshold Monotonicity property, the right hand side is equal to
\mbox{$\bigcap_{F \supseteq \cC(A) \cup \cC(B)} \cC(F)$}, the exact translation
of the right hand side of Equation~\ref{eq:satoh}.

Leaving the exact formulation of the result for later, let us
examine the situation.

The completeness direction seems easier than the soundness one.
Assume \mbox{$\cC : 2^{\cL} \longrightarrow 2^{\cL}$}
satisfies Inclusion, Idempotence, Cautious Monotonicity, 
Conditional Monotonicity, Threshold Monotonicity and property (E).
As in the proof of the main representation result of previous work,
we take \cM\ to be the set of all theories (sets of formulas closed
under \cC) and define the satisfaction relation by:
\mbox{$T \models a$} iff \mbox{$a \in T$}.
We define $f$ by: \mbox{$f(X) = X \cap \widehat{\cC(\overline{X})}$}.
We have shown in the previous proof that $f$ preserves definability
and satisfies Contraction, Coherence and Local Monotonicity and that
\mbox{$\cC(A) = \overline{f(\widehat{A})}$}.
We easily notice that, for any set $X$,
\begin{equation}
\label{eq:XincXbarhat}
f(X) \subseteq f(\widehat{\overline{X}}).
\end{equation}
We already noticed in~\cite{LehLogSemantics:TR} that:
\begin{equation}
\label{eq:Ahatbar}
\overline{\widehat{A}} = \bigcap_{B \supseteq A} \cC(B)
\end{equation}
and that
\begin{equation}
\label{eq:hatCfhat}
f(\widehat{A}) = \widehat{\cC(A)}
\end{equation}

Notice, first, that, by (\ref{eq:Ahatbar}), for any $A$, $B$:
\[
\cC(\bigcap_{F \supseteq A {\rm \ or \ } F \supseteq B} \cC(F)) =
\cC(\overline{\widehat{A}} \cap \overline{\widehat{B}})
\]
and that, by Threshold Monotonicity and by (\ref{eq:Ahatbar}), 
for any $A$, $B$,
\[
\cC(\cC(A) , \cC(B)) = \bigcap_{F \supseteq \cC(A) \cup \cC(B)} \cC(F) =
\overline{{\rm Mod}(\cC(A) \cup \cC(B))}.
\]
Property (E) then becomes:
\[
{\bf (E')} \ \ \ \ \ \cC(\overline{\widehat{A}} \cap \overline{\widehat{B}})
\subseteq \overline{{\rm Mod}(\cC(A) \cup \cC(B))}.
\]
Therefore we have:
\[
{\bf (E'')} \ \ \ \ \ {\rm Mod}(\cC(A) \cup \cC(B)) = 
{\rm Mod}(\overline{{\rm Mod}(\cC(A) \cup \cC(B))}) \subseteq
{\rm Mod}(\cC(\overline{\widehat{A}} \cap \overline{\widehat{B}})).
\]

%

We want to show that Expansion holds true.
Let $X$ and $Y$ be arbitrary sets of models and let us define
\mbox{$A = \overline{X}$} and \mbox{$B = \overline{Y}$}
By definition of $f$:
\[
f(X) \cap f(Y) = X \cap \widehat{\cC(A)} \cap Y \cap 
\widehat{\cC(B)}
\]
and
\[
f(X \cup Y) = (X \cup Y) \cap {\rm Mod}(\cC(\overline{X \cup Y})).
\]
Since \mbox{$X \cap Y \subseteq X \cup Y$}, to prove Expansion,
i.e., \mbox{$f(X) \cap f(Y) \subseteq f(X \cup Y)$},
it is enough to prove that
\[
\widehat{\cC(A)} \cap 
\widehat{\cC(B)} \subseteq 
{\rm Mod}(\cC(\overline{X \cup Y})).
\]
But 
\[
\widehat{\cC(A)} \cap \widehat{\cC(B)} = 
{\rm Mod}(\cC(A) \cup \cC(B))
\]
and
\mbox{$\overline{X \cup Y} = A \cap B$}.
It is therefore enough to show that
\[
{\rm Mod}(\cC(A) \cup \cC(B)) \subseteq {\rm Mod}(\cC(A \cap B)).
\]
By (E'') it is enough to show that:
\[
{\rm Mod}(\cC(\overline{\widehat{A}} \cap \overline{\widehat{B}})) \subseteq
{\rm Mod}(\cC(A \cap B)).
\]
But \mbox{$A = \overline{X} = \overline{\widehat{A}}$} and 
\mbox{$B = \overline{Y} = \overline{\widehat{B}}$}.

We have proved the following.
\begin{theorem}
\label{the:completeness}
If \mbox{$\: \cC : 2^{\cL} \longrightarrow 2^{\cL}$}
satisfies Inclusion, Idempotence, Cautious Monotonicity, 
Conditional Monotonicity, Threshold Monotonicity and property (E),
then, there is a set \cM, a satisfaction relation $\models$
and a definability-preserving choice function $f$ that satisfies
Contraction, Coherence, Local Monotonicity and Expansion such that
\mbox{$\cC(A) = \overline{f(\widehat{A})}$}.
The function $f$ may be chosen to satisfy
\mbox{$f(X) \subseteq f(\widehat{\overline{X}})$}.
\end{theorem}

Let us now turn to the soundness direction.
We assume that the union of two definable set is definable,
i.e.,
\begin{equation}
\label{eq:undef}
\widehat{A} \cup \widehat{B} = {\rm Mod}(\overline{\widehat{A} \cup \widehat{B}}) =
{\rm Mod}(\overline{\widehat{A}} \cap \overline{\widehat{B}}).
\end{equation}
This assumption is satisfied, for example, if the language \cL\ is closed
under a binary connective $\vee$ that behaves semantically as a disjunction:
\begin{equation}
\label{eq:or}
x \models a \vee b \ {\rm iff \ either} \ x \models a \ {\rm or} \ x \models b.
\end{equation}
In this case, if \mbox{$X = \widehat{A}$} and \mbox{$Y = \widehat{B}$},
then \mbox{$X \cup Y = \widehat{A \vee B}$}, where \mbox{$A \vee B$}
is the set of all formulas \mbox{$a \vee b$} for \mbox{$a \in A$} and 
\mbox{$b \in B$}.

We assume that $f$ preserves definability and satisfies Contraction, Coherence,
Local Monotonicity and Expansion.
We also assume
\begin{equation}
\label{eq:incfhatbar}
f(X) \subseteq f(\widehat{\overline{X}}).
\end{equation}
\cC\ is defined by: \mbox{$\cC(A) = \overline{f(\widehat{A})}$}.
We know from previous work that \cC\ satisfies Inclusion, Idempotence,
Cautious Monotonicity, Conditional Monotonicity and Threshold Monotonicity.
Therefore \cC\ also satisfies Cumulativity.
We also know that
\begin{equation}
\label{eq:imp}
\widehat{\cC(A)} = f(\widehat{A})
\end{equation}
and that
\begin{equation}
\label{eq:CnsubC}
\overline{\widehat{A}} \subseteq \bigcap_{F \supseteq A} \cC(F).
\end{equation}
We want to show that property (E) holds:
\[
{\bf (E)} \ \ \ \ \ \cC(\bigcap_{F \supseteq A {\rm \ or \ } F \supseteq B} \cC(F)) \:\subseteq \: \cC(\cC(A) , \cC(B)).
\]
We shall proceed in two stages.
In the first stage we shall show that
\[
\cC(\bigcap_{F \supseteq A {\rm \ or \ } F \supseteq B} \cC(F)) \: \subseteq \:
\cC(\overline{\widehat{A}} \cap \overline{\widehat{B}}).
\]
In the second stage we shall show that 
\[
\cC(\overline{\widehat{A}} \cap \overline{\widehat{B}}) \: \subseteq \:
\cC(\cC(A) , \cC(B)).
\]
Expansion is needed only in the second stage.

Let us deal with the first stage. In fact, we can show equality.
The result relies on the following lemma, asserting what Makinson~\cite{Mak:Handbook}
called Distributivity.
\begin{lemma}
\label{le:dist}
\mbox{$\cC(A) \cap \cC(B) \subseteq \cC(\overline{\widehat{A}} \cap 
\overline{\widehat{B}})$}.
\end{lemma}
\proof
By Equation~\ref{eq:undef},
\[
f({\rm Mod}(\overline{\widehat{A} \cup \widehat{B}})) = f(\widehat{A} \cup \widehat{B}),
\]
but, by Coherence
\[
f(\widehat{A} \cup \widehat{B}) \subseteq f(\widehat{A}) \cup f(\widehat{B}).
\]
We conclude that we have:
\[
f({\rm Mod}(\overline{\widehat{A} \cup \widehat{B}})) \subseteq f(\widehat{A}) \cup f(\widehat{B}).
\]
Therefore, by considering the sets of models of both sides:
\[
\overline{f(\widehat{A})} \cap \overline{f(\widehat{B})} \subseteq 
\overline{f({\rm Mod}(\overline{\widehat{A} \cup \widehat{B}}))}.
\]
By the definition of \cC\ we have
\[
\cC(A) \cap \cC(B) \subseteq \cC(\overline{\widehat{A} \cup \widehat{B}})
= \cC(\overline{\widehat{A}} \cap 
\overline{\widehat{B}}).
\]
\QED
Now, we claim:
\[
\overline{\widehat{A}} \cap \overline{\widehat{B}} \subseteq
\bigcap_{F \supseteq A {\rm \ or \ } F \supseteq B} \cC(F)) \subseteq
\cC(A) \cap \cC(B) \subseteq
\cC(\overline{\widehat{A}} \cap \overline{\widehat{B}}).
\]
The first inclusion follows from Equation~\ref{eq:CnsubC},
the second inclusion is obvious and the third one follows from Lemma~\ref{le:dist}.
But we know that \cC\ is cumulative and therefore
\[
\cC(\bigcap_{F \supseteq A {\rm \ or \ } F \supseteq B} \cC(F)) \: = \:
\cC(\overline{\widehat{A}} \cap \overline{\widehat{B}}).
\]

For the second stage, notice that, by Expansion
\[
f(\widehat{A}) \cap f(\widehat{B}) \subseteq f(\widehat{A} \cup \widehat{B}).
\]
By Contraction
\[
f(f(\widehat{A}) \cap f(\widehat{B})) \subseteq f(\widehat{A}) \cap f(\widehat{B})
\]
and by Equation~\ref{eq:incfhatbar},
\[
f(\widehat{A} \cup \widehat{B}) \subseteq 
f({\rm Mod}(\overline{\widehat{A} \cup \widehat{B}})),
\]
therefore we have
\[
f(f(\widehat{A}) \cap f(\widehat{B})) \subseteq
f({\rm Mod}(\overline{\widehat{A} \cup \widehat{B}})).
\]
By considering the sets of formulas defined by both sides we have:
\[
\overline{f({\rm Mod}(\overline{\widehat{A} \cup \widehat{B}}))} \subseteq
\overline{f(f(\widehat{A}) \cap f(\widehat{B}))}.
\]
But, by Equation~\ref{eq:imp},
\[
f(f(\widehat{A}) \cap f(\widehat{B})) = f(\widehat{\cC(A)} \cap \widehat{\cC(B)}) =
f({\rm Mod}(\cC(A) \cup \cC(B))).
\]
By the definition of \cC\ we have:
\[
\cC(\overline{\widehat{A} \cup \widehat{B}})) \subseteq
\cC(\cC(A) , \cC(B)),
\]
which concludes the second stage of our derivation of (E).
\QED
We summarize this soundness result.
\begin{theorem}
\label{the:sound}
If \cM is a set of models and 
\mbox{$\models \: \subseteq \cM \times \cL$} is such that the
union of any two definable sets of models is definable, and if
\mbox{$f : \cM \longrightarrow \cM$} is definability-preserving and
satisfies Contraction, Coherence, Local Monotonicity, 
Expansion and \mbox{$f(X) \subseteq f(\widehat{\overline{X}})$}, 
then the operation defined by:
\mbox{$\cC(A) = \overline{f(\widehat{A}})$} satisfies 
Inclusion, Idempotence, Cautious Monotonicity, 
Conditional Monotonicity, Threshold Monotonicity and property (E).
\end{theorem}

One would like to prove a stronger and more elegant result.
Assume that the language \cL\ is closed under a binary connective $\vee$.
One may show, as in Theorem~\ref{the:sound} that:
if \cM is a set of models and 
\mbox{$\models \: \subseteq \cM \times \cL$} is such that 
\mbox{$m \models a \vee b$} iff \mbox{$m \models a$} or
\mbox{$m \models b$}, and if
\mbox{$f : \cM \longrightarrow \cM$} is definability-preserving and
satisfies Contraction, Coherence, Local Monotonicity, 
Expansion and \mbox{$f(X) \subseteq f(\widehat{\overline{X}})$}, 
then the operation defined by:
\mbox{$\cC(A) = \overline{f(\widehat{A}})$} satisfies 
Inclusion, Idempotence, Cautious Monotonicity, 
Conditional Monotonicity, Threshold Monotonicity, property (E),
\begin{equation}
\label{eq:orleftintro}
\cC(A , a) \cap \cC(A , b) \subseteq \cC(A , a \vee b),
\end{equation}
and
\begin{equation}
\label{eq:orrightintro}
a \in \cC(A) \: \Rightarrow \: a \vee b \in \cC(A) , \:
b \in \cC(A) \: \Rightarrow \: a \vee b \in \cC(A). 
\end{equation} 

One would like to strengthen Theorem~\ref{the:completeness}
by claiming:
if \mbox{$\cC : 2^{\cL} \longrightarrow 2^{\cL}$}
satisfies Inclusion, Idempotence, Cautious Monotonicity, 
Conditional Monotonicity, Threshold Monotonicity, property (E),
(\ref{eq:orleftintro}) and (\ref{eq:orrightintro}),
then, there exists a set \cM, a satisfaction relation $\models$
such that 
\mbox{$m \models a \vee b$} iff \mbox{$m \models a$} or
\mbox{$m \models b$},
and a definability-preserving choice function $f$ that satisfies
Contraction, Coherence, Local Monotonicity, Expansion 
and \mbox{$f(X) \subseteq f(\widehat{\overline{X}})$}
such that
\mbox{$\cC(A) = \overline{f(\widehat{A})}$}.

Unfortunately this result does not seem to be within reach,
because some compactness assumption seems to be needed to show
that if $T$ is a theory that does not contain $a$, there is
a $\vee$-complete theory $T' \supseteq T$ that does not contain $a$,
where $\vee-complete$ means: \mbox{$a \vee b \in T$} implies either
\mbox{$a \in T$} or \mbox{$b \in T$}.

A more elegant equivalence result, on the model of Theorem 5 
of~\cite{LehLogSemantics:TR} may be obtained if one assumes
the existence of proper negation and disjunction.
We need to consider the following syntactic and semantic properties.
\[
{\bf Weak \ Compactness} \ \ \ \ \cC(A) = \cL \: \Rightarrow \:
\exists {\rm \ a \ finite \ } B \subseteqf A \ {\rm such \ that} \ 
\cC(B) = \cL.
\]
\begin{equation}
\label{eq:negleftintro}
\cC(A , a , \neg a) = \cL
\end{equation}
\begin{equation}
\label{eq:negleftelim}
\cC(A , \neg a) = \cL \Rightarrow a \in \cC(A)
\end{equation}
\begin{equation}
\label{eq:neg}
x \models \neg a \ {\rm iff} \ x \not \models a .
\end{equation}

\begin{theorem}
\label{the:wrap}
Assume \cL\ is closed under a unary connective ($\neg$) and
a binary connective ($\vee$) and that 
\mbox{$\cC : 2^{\cL} \longrightarrow 2^{\cL}$}
satisfies {\bf Weak Compactness}, (\ref{eq:negleftintro}),
(\ref{eq:negleftelim}),(\ref{eq:orleftintro}), (\ref{eq:orrightintro}),
Inclusion, Idempotence, Cautious Monotonicity, Conditional Monotonicity
and Threshold Monotonicity.
The two following conditions are equivalent:
\begin{enumerate}
\item \label{e}
\cC\ satisfies property (E), and
\item \label{f}
there is a set \cM, a satisfaction relation $\models$ that
satisfies~(\ref{eq:neg}), (\ref{eq:or}) and
a definability-preserving choice function $f$ that satisfies
Contraction, Coherence, Local Monotonicity, Expansion and~(\ref{eq:incfhatbar})
such that
\mbox{$\cC(A) = \overline{f(\widehat{A})}$}, for any \mbox{$A \subseteq \cL$}.
\end{enumerate}
\end{theorem}
\proof
Assume~\ref{e}. By Theorem 5 of~\cite{LehLogSemantics:TR} and
Theorem~\ref{the:completeness}, \ref{f} follows.
Assume~\ref{f}. The union of two definable sets of models is definable,
as explained following (\ref{eq:or}).
Theorem~\ref{the:sound} implies that property (E) holds.
\QED

\bibliographystyle{plain}

\end{document}